\documentclass{article}

    \usepackage[final]{neurips_2024}

\usepackage[utf8]{inputenc} 
\usepackage[T1]{fontenc}    
\usepackage{hyperref}       
\usepackage{url}            
\usepackage{booktabs}       
\usepackage{amsfonts}       
\usepackage{nicefrac}       
\usepackage{microtype}      
\usepackage{xcolor}         
\usepackage{graphicx}
\usepackage{subcaption}
\usepackage{amsmath}
\usepackage{authblk}
\usepackage{float}
\usepackage{cleveref}

\title{Robotic Assistant: Completing Collaborative Tasks with Dexterous Vision-Language-Action Models}
\author[1]{Boshi An, Chenyu Yang, Robert Katzschmann}
\affil[1]{Work done at Soft Robotics Lab, ETHz, Switzerland}
\date{Feb 2025}

\begin{document}

\maketitle

\section{Introduction}

Robots are autonomous mechanical systems designed to assist humans with complex tasks. In many scenarios, effective human-robot collaboration is essential, whether through physical assistance or more nuanced forms of interaction. However, building robots that can collaborate naturally and intuitively with humans remains a significant challenge. Directly training control policies on specific tasks often leads to overfitting and fails to capture high-level task semantics or human intent.

Large language models (LLMs) have recently demonstrated impressive capabilities in reasoning, generalization, and multimodal understanding, making them promising candidates for enabling more flexible robotic behaviors. Yet, directly applying LLMs to real-world collaborative robotics remains impractical for two key reasons: (1) LLMs lack the mechanisms to bridge the gap between abstract reasoning and low-level control, and (2) they rely heavily on explicit language prompting, which introduces latency and inefficiency in real-time interactions.

To enable smoother and more intuitive collaboration, we envision a policy that minimizes the need for language-based prompting and instead infers human intent directly from motion cues — enabling a robot to act through \textbf{tacit understanding}.

In this project, we propose a novel approach that fine-tunes pre-trained vision-language-action (VLA) models for collaborative tasks. We introduce several key modifications to improve adaptation to real-world human-robot interaction: leveraging pre-trained visual encoders, incorporating human pose priors, and re-designing the model's action space. Our approach enhances the robot’s ability to perceive, interpret, and respond to human behaviors in a context-aware and data-efficient manner. Real-world evaluations demonstrate the effectiveness of the proposed method.

\section{Related Work}




\subsection{Human-Robot Interaction}

Human-robot interaction (HRI) is a longstanding area of research aimed at improving the ways in which robots assist and collaborate with humans. Prior work has explored a variety of methods to enhance robot responsiveness, intention understanding, and physical cooperation. For example, Roveda et al.~\cite{roveda2019assisting} employed fuzzy controllers to support humans in industrial settings. Yan et al.~\cite{yan2019human} used long short-term memory (LSTM) networks for intention recognition in human-robot interaction. Similarly, Zhang et al.~\cite{zhang2020recurrent} applied recurrent models to predict human motion during assembly tasks to facilitate handovers. More recently, Wojtak et al.~\cite{wojtak2021neural} proposed using neural fields for learning object handover behaviors, while Ji et al.~\cite{ji2024foundation} and Wang et al.~\cite{wang2024lami} explored foundation model-based approaches for collaborative assembly and tabletop interaction, respectively. However, these methods often depend on handcrafted robotic APIs and suffer from high inference latency, limiting their real-time applicability.

In contrast, we propose a system that enables real-time, smooth human-robot collaboration by directly generating robot actions from multimodal observations, without relying on predefined action schemas.





\subsection{Learning from Demonstrations}

Learning from demonstrations (LfD), also known as imitation learning, is a widely adopted paradigm in robotic learning~\cite{zare2024survey}. By mimicking human behavior, robots can acquire complex skills without requiring manually designed reward functions. Classical approaches include Behavior Cloning (BC), which maximizes the likelihood of expert actions given observed states, and Inverse Reinforcement Learning (IRL)~\cite{arora2021survey}, which infers the underlying reward function from demonstrations. DAgger~\cite{ross2011reduction} addresses distributional shift by iteratively querying the expert in an online setting.

To improve data efficiency and handle imperfect demonstrations, more recent methods incorporate probabilistic and generative modeling. Huang et al.~\cite{huang2018generalized} proposed a Gaussian Mixture Model (GMM)-based framework for few-shot learning in long-horizon tasks, while Bütepage et al.~\cite{butepage2020imitating} used generative models for imitation in human-robot interaction scenarios.

The emergence of large-scale robotic datasets~\cite{brohan2022rt,o2024open} has enabled the development of generalist policies trained with simple imitation objectives. These datasets support scaling imitation learning to diverse tasks and environments.





\subsection{Vision-Language-Action (VLA) Models}

Recent advances in large language models (LLMs)~\cite{brown2020language,achiam2023gpt} have demonstrated strong capabilities in reasoning, abstraction, and multimodal alignment. This has motivated efforts to apply LLMs to robotics, where they could bridge perception and action through natural language.

Preliminary works such as Text2Motion~\cite{lin2023text2motion} and VoxPoser~\cite{huang2023voxposer} have explored this direction. Building on large-scale multimodal datasets and vision-language pretraining~\cite{anil2023palm,liu2023visual,liu2024improved}, researchers have introduced VLA models that process visual and linguistic inputs to directly generate tokenized robot actions~\cite{driess2023palm,kim2024openvla,team2024octo}. These models are trained using next-token prediction over sequences of multimodal inputs and demonstrations.

VLA models exhibit strong generalization and compositionality, allowing them to handle open-ended, unstructured tasks. They can also be adapted to specific domains via fine-tuning, making them a promising foundation for learning collaborative robot behaviors from modest data.

\section{Method}

\subsection{Robotic System}

\begin{figure}
    \centering
    \includegraphics[width=0.5\linewidth]{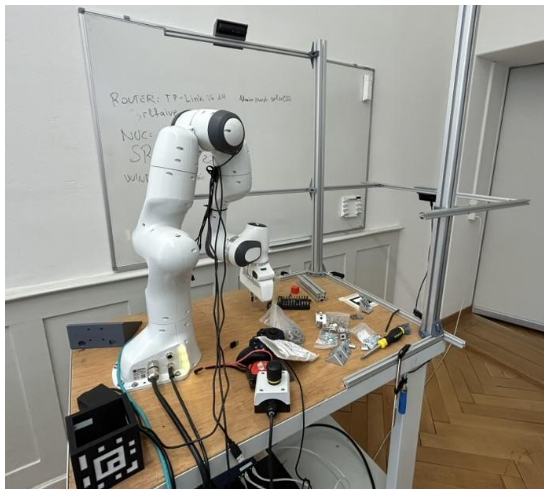}
    \caption{Robotic system}
    \label{fig:robotic-system}
\end{figure}

We use a Mimic hand mounted on a Franka Panda robotic arm. Two Mimic hand cameras and two external cameras are used to capture visual input. The full system is shown in~\cref{fig:robotic-system}.

A teleoperation system is also part of the robotic system. We use Rokoko mocap gloves to capture the absolute position, rotation and finger pose of human hands to allow teleoperation of the robot. The mocap data is mapped to the mimic hand and Franka Panda arm via a motion retargeting system.

\subsection{Data Collection Pipeline}





\subsubsection{Collection}
\label{collection}

To collect training data for model training, we designed a collaborative data collection pipeline involving two human participants: the \textbf{teleoperator} and the \textbf{collaborator}. The teleoperator controls the robot by wearing Rokoko motion capture gloves, which record the absolute position, orientation, and finger articulation of their hands. This motion data is then retargeted onto the robotic system to enable teleoperation.

\begin{figure}[htbp]
  \centering
  \begin{subfigure}[b]{0.45\textwidth}
    \includegraphics[width=\textwidth]{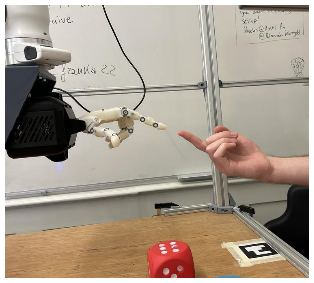}
    \caption{Collaborator}
    \label{fig:first}
  \end{subfigure}
  \hfill
  \begin{subfigure}[b]{0.45\textwidth}
    \includegraphics[width=\textwidth]{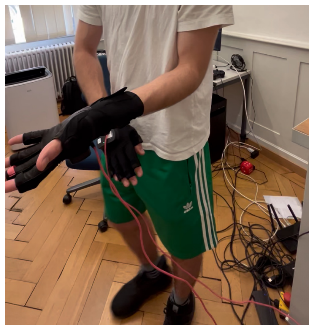}
    \caption{Teleoperator}
    \label{fig:second}
  \end{subfigure}
  \caption{The individuals involved in data collection.}
  \label{fig:data-collection-individuals}
\end{figure}

The collaborator interacts with the robot in a shared workspace, enabling natural human-robot interaction to unfold. The two roles are illustrated in~\cref{fig:data-collection-individuals}.

During each trial, the teleoperator controls the robot to perform collaborative tasks with the human collaborator. Throughout the trial, raw robot states, camera inputs, and action commands are streamed and recorded in HDF5 format.






\subsubsection{Post-processing}
\label{post-processing}

After data collection, the raw streams are post-processed to construct structured datasets suitable for model training. This includes synchronizing all sensory and control data and extracting snapshot frames at a fixed frequency. For this project, we used a sampling rate of 10 Hz.

In addition to the synchronized sensory data, we augment each frame with a text prompt and several auxiliary labels. The text prompt encodes the intended command for the robot during the task (e.g., "pick up the red cube"). The auxiliary labels provide extra supervision to guide the model’s understanding of human intent. Specifically, we include:

\begin{itemize}
    \item The 3D hand pose of the human collaborator, estimated using the Mediapipe hand pose detector~\cite{mediapipe2024}
    \item The index of the target object the human intends to interact with
\end{itemize}

These labels serve to improve the model’s ability to interpret human motion and disambiguate collaborative goals.

\begin{figure}
    \centering
    \includegraphics[width=0.9\linewidth]{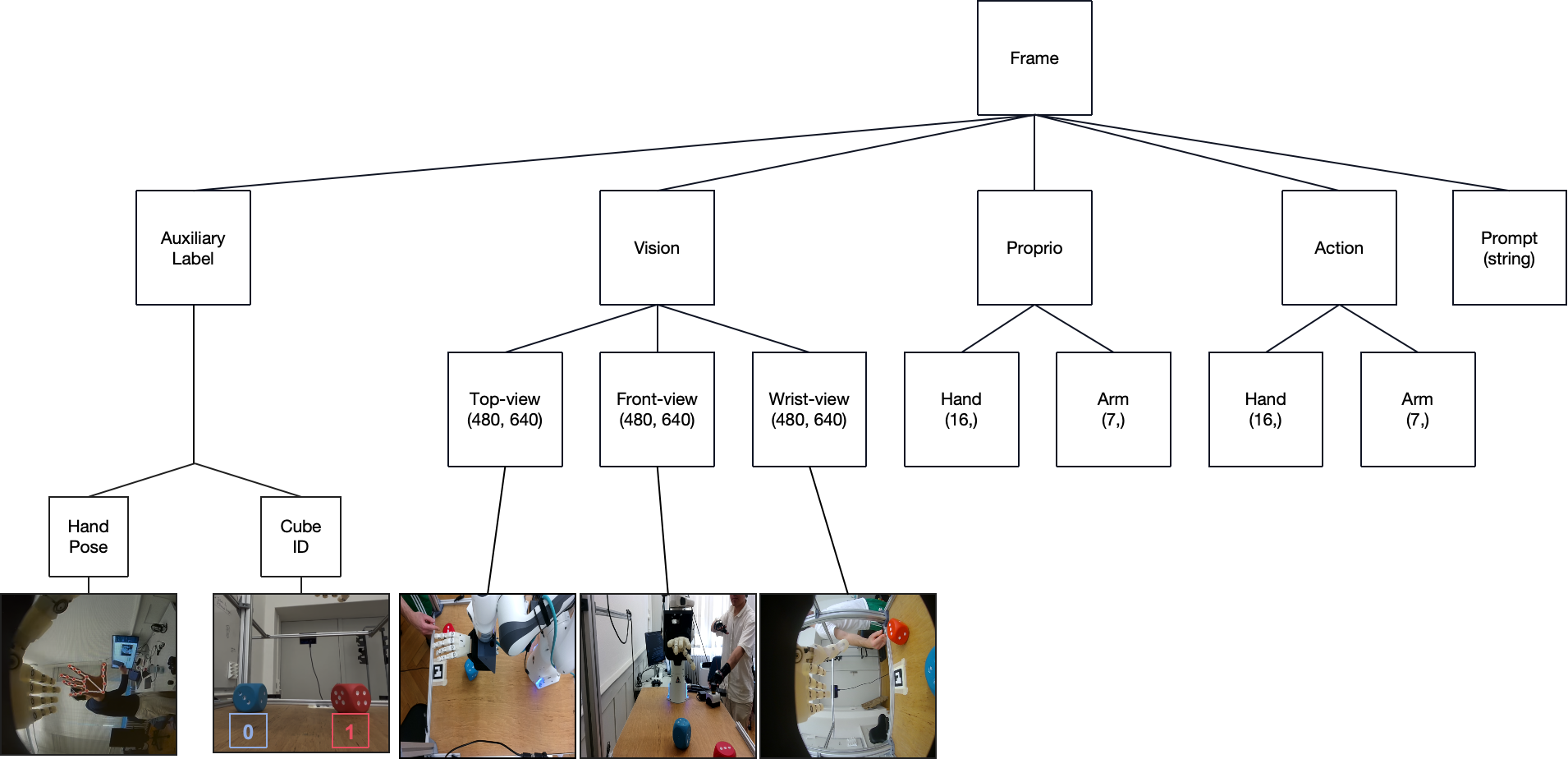}
    \caption{The final composition of the synchronized dataset.}
    \label{fig:dataset_composition}
\end{figure}

The final structure of the synchronized dataset is illustrated in~\cref{fig:dataset_composition}.

\subsection{Task Design}

We design two toy tasks for this project: "\textit{pick up cube}" and "\textit{pass cube}". These tasks were carefully selected because they are illustrative of core capabilities required for human-robot collaboration. Specifically:

\begin{enumerate}
    \item They demonstrate the robot's ability to assist a human physically, through object manipulation and transfer.
    \item They can be composed into a longer sequence — first picking up an object indicated by the human, then passing it back — showcasing the model’s ability to execute long-horizon, goal-directed behavior.
    \item They require the robot to interpret human body language rather than relying on explicit natural language instructions, aligning with our goal of enabling tacit understanding.
\end{enumerate}

The "\textit{pick up cube}" task involves two cubes placed on a table — one red and one blue. The human collaborator points to one cube, and the robot must infer the intention and pick up the designated object.

The "\textit{pass cube}" task begins with the robot already holding a cube. The robot is required to pass the object to the human collaborator and release it appropriately.

We collected 60 trajectories for each cube in the pick-up task (120 total), and 200 trajectories for the red cube and 60 for the blue cube in the pass task.





\subsection{Vision-Language-Action Model}

Our approach builds upon Open-VLA~\cite{kim2024openvla}, a recently proposed vision-language-action model. For visual perception, Open-VLA incorporates pre-trained encoders from SigLIP~\cite{zhai2023sigmoid} and DINOv2~\cite{oquab2023dinov2}. Language inputs are processed using a pre-trained LLaMA2-7B model~\cite{touvron2023llama}. These components are integrated into a unified multimodal transformer that fuses visual, linguistic, and proprioceptive inputs to generate robot actions.

\begin{figure}[ht]
    \centering
    \includegraphics[width=0.9\linewidth]{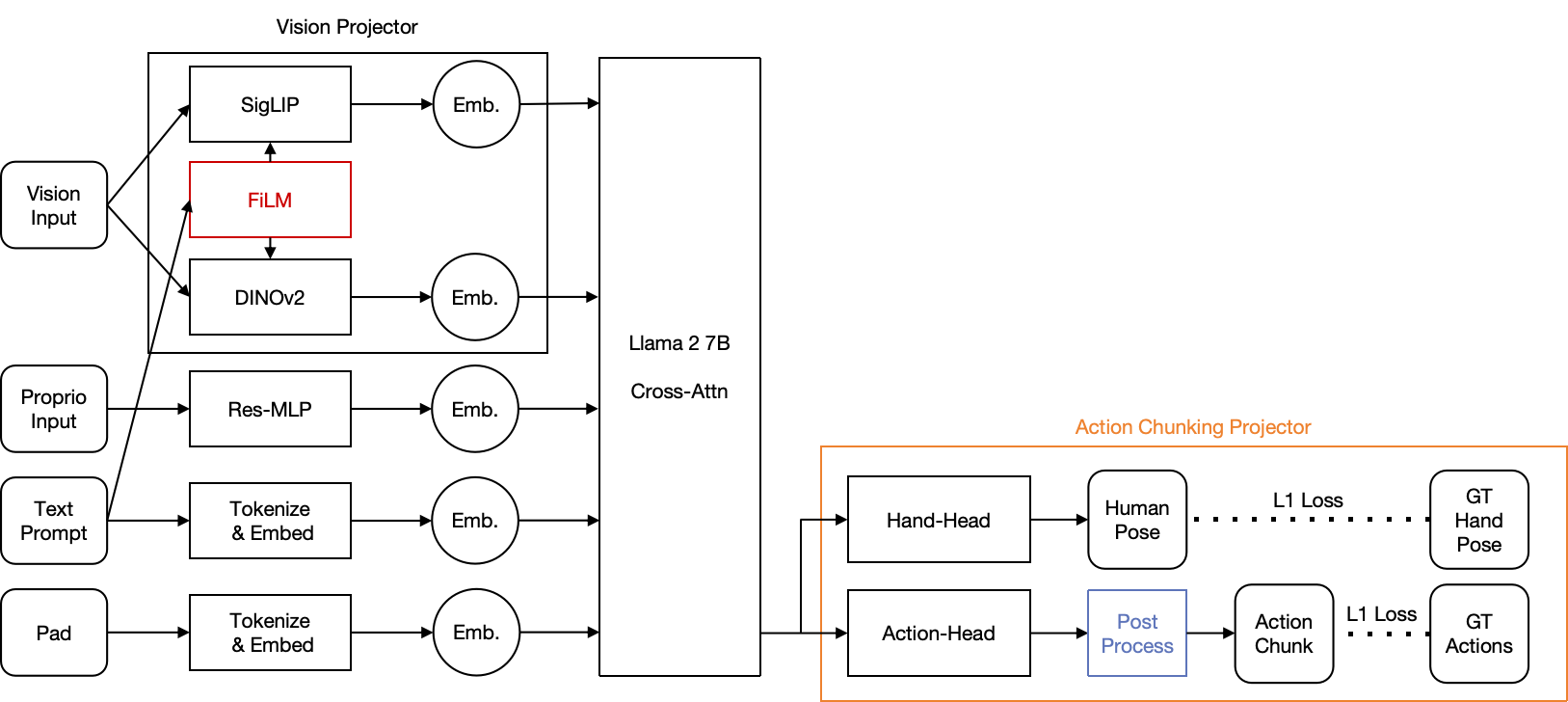}
    \caption{The modified model structure. Red block represents the FiLM layers added to vision encoders, orange block represents the modified action chunking projector, blue block represents the modified action post-processing module.}
    \label{fig:model}
\end{figure}

To better adapt Open-VLA to collaborative settings, we introduce several key modifications, illustrated in~\cref{fig:model}, and analyzed in subsequent sections:

\begin{enumerate}
    \item \textbf{FiLM conditioning}~\cite{perez2018film}: We insert FiLM layers into both vision encoders to improve cross-modal conditioning from text.
    \item \textbf{Auxiliary intention loss}: We add an auxiliary prediction head to explicitly learn human intention by regressing collaborator hand pose.
    \item \textbf{Action post-processing}: We constrain action predictions to a more compact and structured subspace, improving stability and learning efficiency.
    \item \textbf{Directional loss}: We apply a directional loss on end-effector pose that emphasizes directional alignment while downweighting magnitude.
\end{enumerate}

These modifications collectively improve the model's ability to interpret human cues and generate responsive, context-aware robot behavior in collaborative tasks.

\subsection{FiLM Conditioning}

Feature-wise Linear Modulation (FiLM)~\cite{perez2018film} is a technique for conditioning a vision encoder on additional inputs, typically text. FiLM layers apply affine transformations to feature maps, where the scale and bias are functions of the conditioning input. This enables the model to dynamically adjust visual representations based on linguistic context.

In the context of VLA models, FiLM layers allow the vision backbone to better align visual perception with task-specific language prompts. We incorporate FiLM conditioning into both vision encoders (SigLIP and DINOv2) and evaluate its impact on task performance in collaborative settings.





\subsection{Auxiliary Loss}

To enhance the model’s understanding of human intent, we introduce human pose priors into training. A straightforward approach would be to extract pose-related features and feed them into the model via cross-attention. However, this method does not scale well: as the number of tasks and priors (e.g., grasp points, object bounding boxes) increases, it would require designing and maintaining multiple feature extractors.

Instead, we adopt an auxiliary loss formulation that encourages the model to implicitly learn human intention cues. Specifically, we add an auxiliary prediction head—referred to as the \textit{hand head}—in parallel with the \textit{action head}. This head receives the same model input and is trained to predict: (1) the 2D hand pose of the collaborator in each camera view, and (2) the color of the target cube. Hand pose annotations are extracted using MediaPipe~\cite{mediapipe2024}, and the target object label is derived from task metadata.

The auxiliary loss is defined as the L2 distance between the predicted and ground-truth labels. During inference, the \textit{hand head} is disabled, as it does not contribute to action generation.








\subsection{Action Post-processing}

The original action space—comprising 3D position, 4D rotation (quaternion), and 16 joint positions—totals 23 dimensions. However, the underlying structure of valid actions likely lies on a lower-dimensional manifold, making it difficult for the model to learn effectively in the raw space.

To address this, we reformulate the action space so that the model predicts actions in a compact, transformed space, which are then mapped back to the original representation via post-processing. This process consists of three stages:

\begin{itemize}
    \item \textbf{Position}: Let $p$ denote the current end-effector position, and $a_p'$ the model's predicted delta. The final position command is computed as $a_p = p + a_p'$.
    
    \item \textbf{Rotation}: Let $q$ be the current end-effector rotation in quaternion form, and $d = (\omega, x, y, z)$ be the a delta quaternion. The output rotation is computed as $a_r = q \cdot d$. The model predicts in the rotation vector form: $a_r' = \frac{(x, y, z)}{\sqrt{1 - \omega^2}}$.
    
    \item \textbf{Hand joints}: We apply PCA to the 16-dimensional hand joint states in the training data and retain the top principal components. During inference, the model predicts in this low-dimensional PCA space, and the full joint configuration is reconstructed via inverse PCA.
\end{itemize}





\subsection{Directional Loss}

To improve the stability and relevance of end-effector motion, we design a \textbf{directional loss} that emphasizes the direction of movement rather than its magnitude. Let $x$ denote the predicted delta pose, and $y$ the ground-truth delta pose. We decompose $x$ into two orthogonal components: $x_\parallel$ (parallel to $y$) and $x_\perp$ (orthogonal to $y$).

The directional loss is defined as:
\[
\mathcal{L}_{\text{dir}} = r \cdot \frac{\|x_\parallel\|_2}{\|y\|_2 + r} + \|x_\perp\|_2,
\]
where $r < 1$ is a scaling factor. When $\|y\|_2 \gg r$, the loss emphasizes directional alignment; when $\|y\|_2 \ll r$, it reduces to a standard L2 loss.

We evaluate this loss function's effect on model performance in later sections.









\subsection{Training Pipeline}

Our training pipeline is based on OpenVLA-OFT~\cite{kim2025fine}, with significant modifications to support collaborative learning. Upon model initialization, the entire network is cast to \texttt{bfloat16} to reduce memory usage, and LoRA adapters are injected into all linear layers to enable efficient fine-tuning. FiLM adapters are injected to both vision encoders depending on whether the option is enabled.

The training data is collected using the procedure described in~\cref{collection,post-processing}. Each dataset file contains a single demonstration trajectory. During training, frames are sampled randomly from these trajectories by the data loader. Each sampled frame is pre-processed into model-ready tensors in the training processor, then assembled into mini-batches by the training collator.

The model receives the processed inputs and computes the predicted outputs. A composite loss—consisting of action loss and auxiliary loss—is computed and optimized using the Adam optimizer~\cite{adam2014method}.

\begin{figure}[H]
    \centering
    \includegraphics[width=0.9\linewidth]{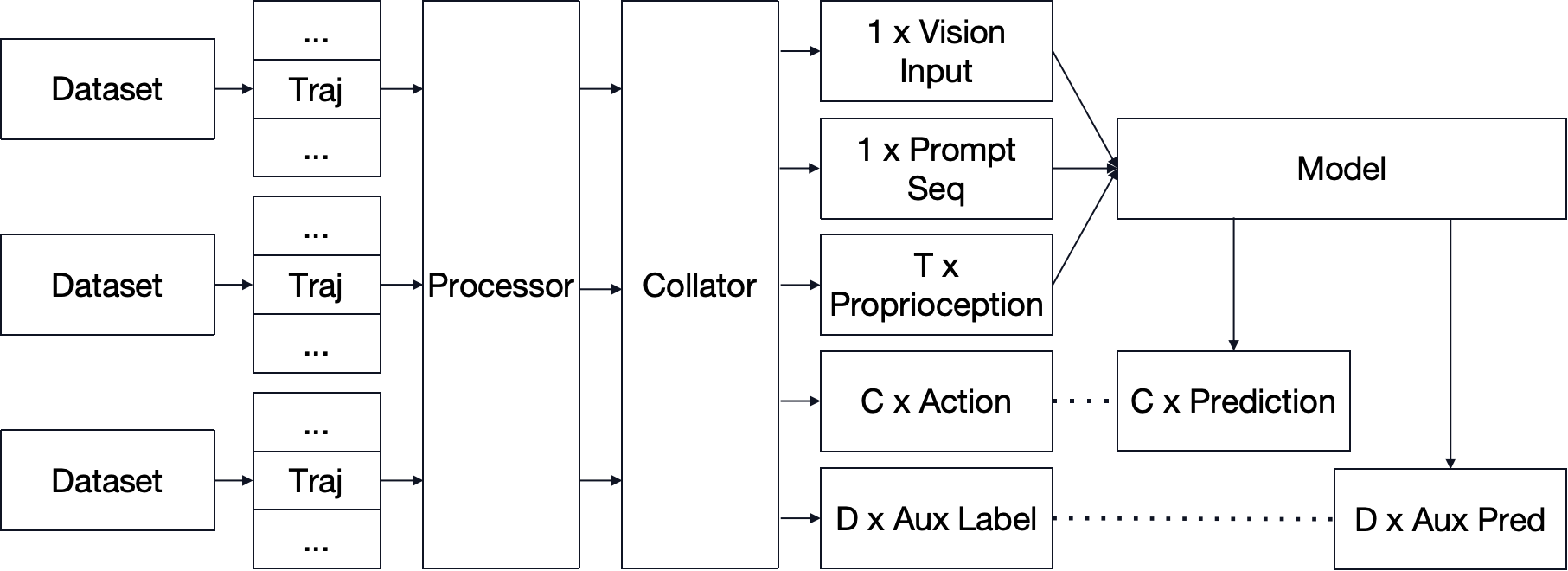}
    \caption{Overview of the training pipeline. The training is carried out with data distribution on a 4 GPU computation node.}
    \label{fig:training-pipeline}
\end{figure}

The full training pipeline is illustrated in~\cref{fig:training-pipeline}.

We carefully tuned the hyperparameters to ensure stable and efficient data-distributed training on a 4×H100 GPU cluster. The final configuration is summarized below:

\begin{itemize}
    \item \textbf{Batch size}: 6 samples/GPU × 4 GPUs = 24 samples/iteration
    \item \textbf{Epochs}: 20
    \item \textbf{Learning rate}: 3e-4
    \item \textbf{LoRA rank}: 32
    \item \textbf{Action chunk size}: 16
    \item \textbf{Proprioception history length}: 2
    \item \textbf{Vision history length}: 1
    \item \textbf{Head hidden size}: 1024
    \item \textbf{LLM hidden size}: 4096
    \item \textbf{Average training runtime}: ~12 hours
\end{itemize}

\subsection{Inference Pipeline}

\label{inference_pipeline}

During inference, the system must stream real-time observations from the robot and cameras to the model, which in turn outputs action predictions with minimal latency.

We designed a dedicated robot interface to manage low-level hardware communication and forward sensory data to a model host. The model host runs the fine-tuned collaborative VLA model in inference mode, processes the incoming data, and returns the predicted robot actions. To support long-horizon tasks, we integrate a rule-based high-level planner that dynamically generates text prompts, allowing the model to chain multiple primitive behaviors into goal-directed sequences.

In our demonstration, we combined the "pick up cube" and "pass cube" tasks into a single long-horizon pipeline. The high-level planner monitors the vertical position of the robotic hand, and once a threshold height is exceeded—indicating that the pickup is complete—it automatically switches the text prompt from a pickup command to a passing command. This is aligned with the way demonstrations were collected: the teleoperator always lifts the robot hand after picking up a cube, providing a reliable signal for transition.

\begin{figure}
    \centering
    \includegraphics[width=0.9\linewidth]{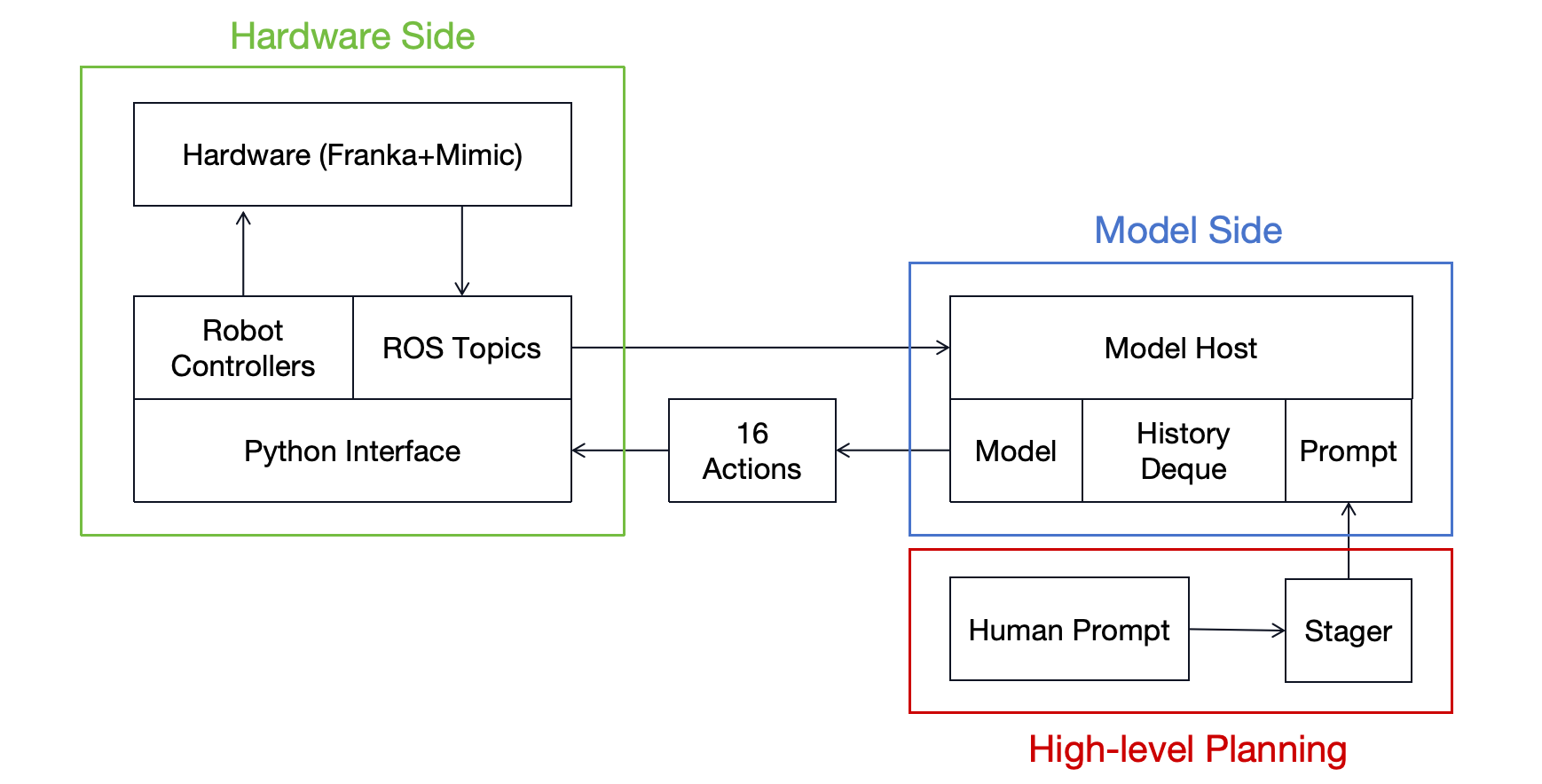}
    \caption{Inference pipeline. The full pipeline consists of a hardware side interface (green), a model side interface (blue) and a high-level planner (red).}
    \label{fig:inference-pipeline}
\end{figure}

The overall inference pipeline is illustrated in~\cref{fig:inference-pipeline}. When executed on a laptop equipped with an NVIDIA RTX 4090 GPU, the end-to-end latency (including model inference and action mapping) was approximately 0.3 seconds. While this was found to be marginally acceptable by human collaborators, further latency reductions remain an important direction for future improvement.

\section{Analysis}

\subsection{Action Space Analysis}

The motivation behind action post-processing is based on the assumption that the true action space lies on a low-dimensional manifold embedded in a high-dimensional space. Without explicitly modeling this structure, the model may struggle to learn meaningful mappings. By reducing the dimensionality, the action manifold can be transformed into a more compact and convex representation, facilitating more efficient learning.

\begin{figure}[htbp]
  \centering
  \begin{subfigure}[b]{0.49\textwidth}
    \includegraphics[width=\textwidth]{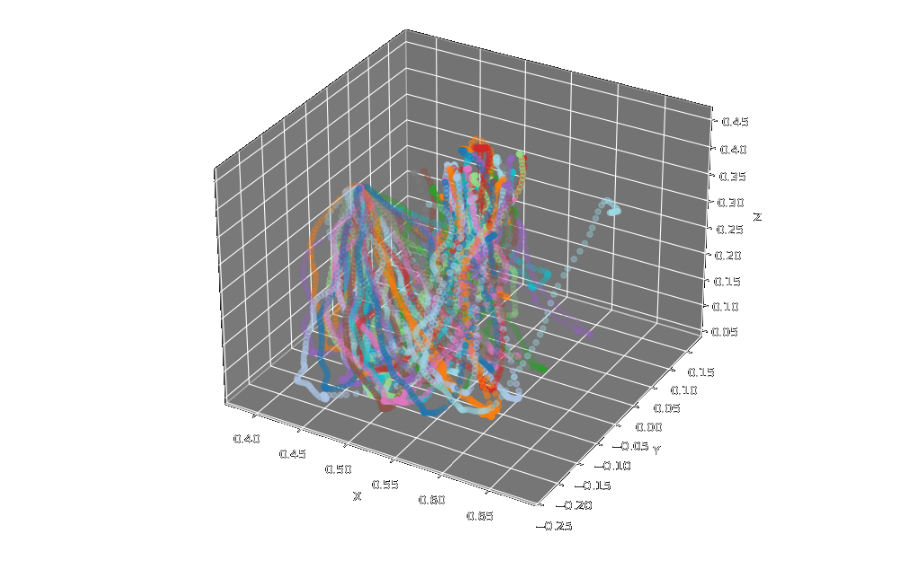}
    \caption{Original action distribution}
    \label{fig:action_pos_first}
  \end{subfigure}
  \hfill
  \begin{subfigure}[b]{0.49\textwidth}
    \includegraphics[width=\textwidth]{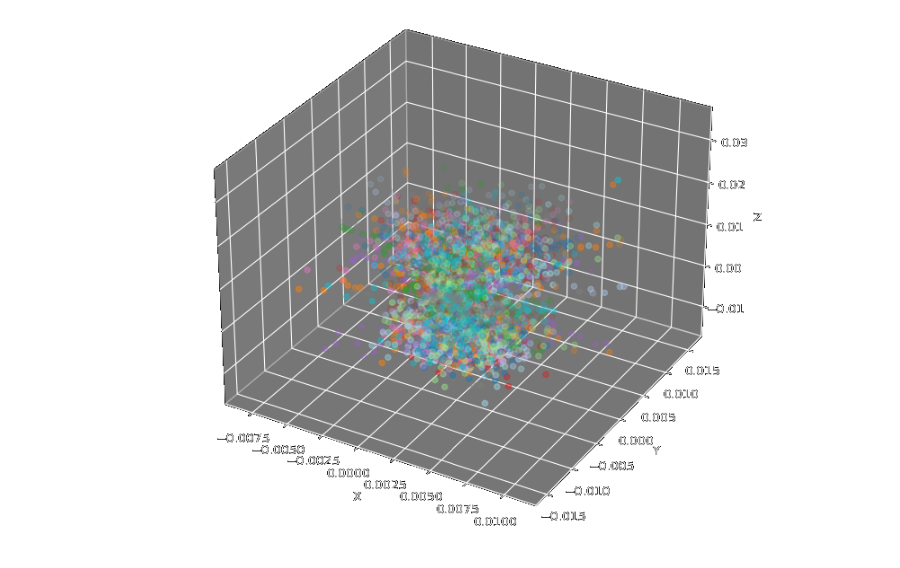}
    \caption{Delta action distribution}
    \label{fig:action_pos_second}
  \end{subfigure}
  \caption{Differentiating the action sequence results in a smoother and more normally distributed action space.}
  \label{fig:action_space_pos}
\end{figure}

We first analyze the action subspace related to end-effector pose. As shown in~\cref{fig:action_pos_first}, the distribution of the raw $xyz$ position components across trajectories is highly non-convex. However, when we differentiate the action sequence—i.e., consider relative rather than absolute motion—the resulting delta poses exhibit a much smoother and near-Gaussian distribution~\cref{fig:action_pos_second}.

\begin{figure}[htbp]
    \centering
    \includegraphics[width=0.9\linewidth]{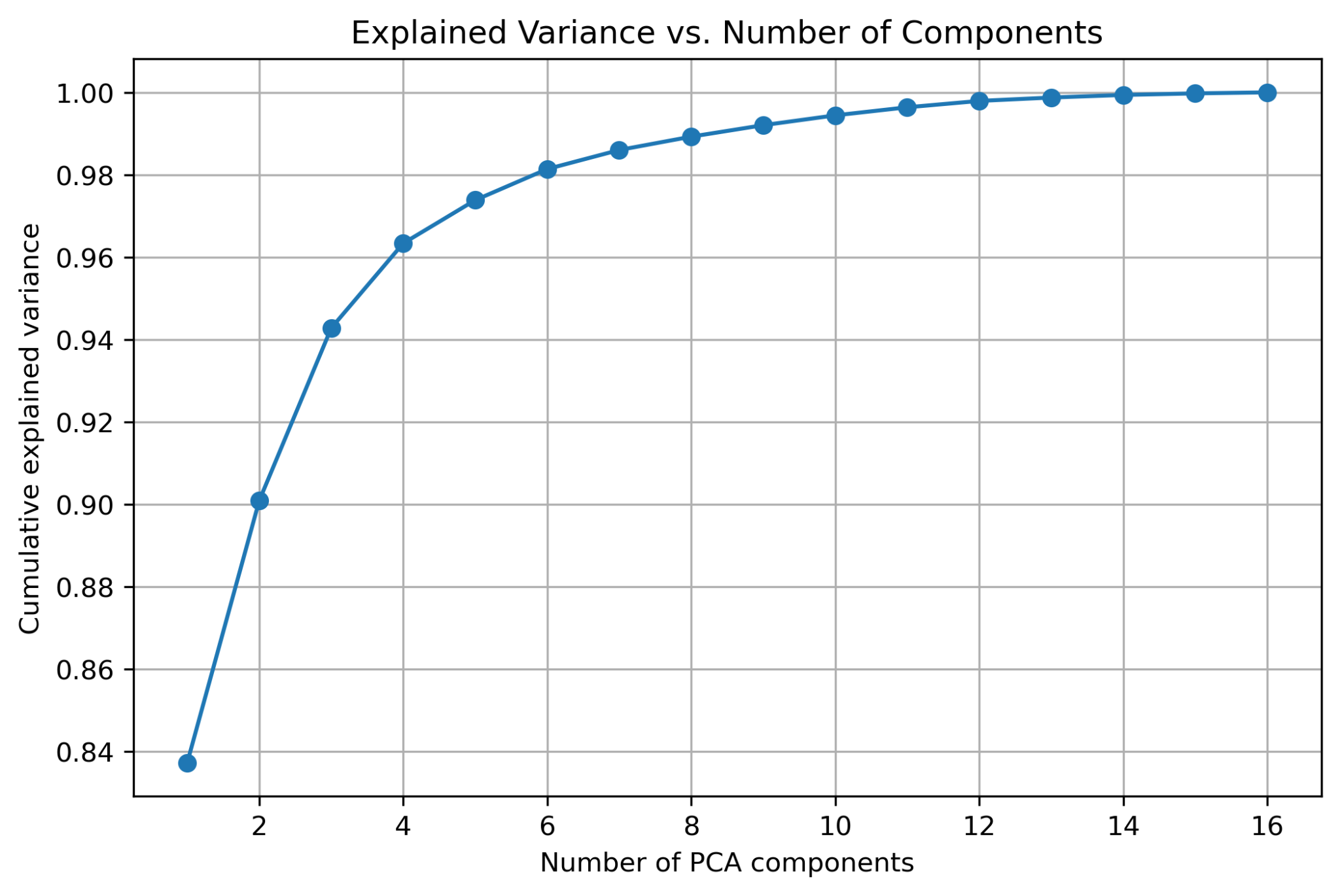}
    \caption{PCA analysis of hand joint states. Four principal components explain 96\% of the variance.}
    \label{fig:hand-pca}
\end{figure}

Next, we examine the hand joint subspace, which has 16 dimensions. We hypothesize that despite this high dimensionality, the actual configuration space is low-dimensional. To test this, we perform principal component analysis (PCA) on all hand joint states in the training set. The results, shown in~\cref{fig:hand-pca}, reveal that just four principal components account for 96\% of the total variance. This suggests that PCA-reduced components can be effectively used as the action representation, replacing the original high-dimensional hand joint space.

\subsection{Ablation Studies}

\begin{figure}
    \centering
    \includegraphics[width=0.9\linewidth]{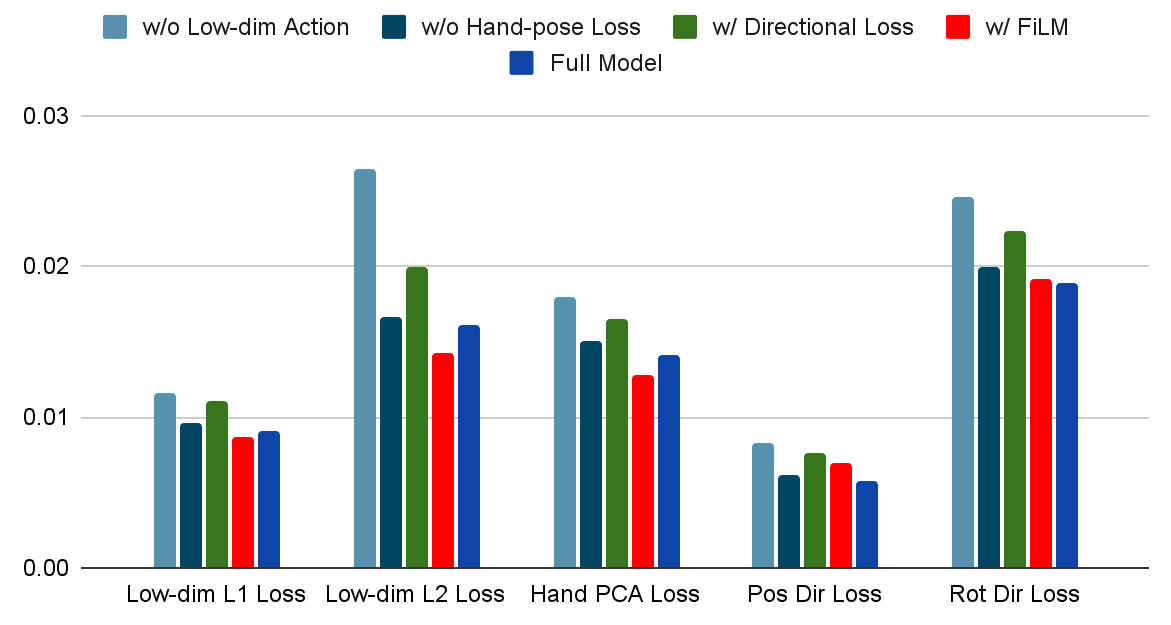}
    \caption{Ablation study results. The “Full Model” includes action post-processing and hand-pose auxiliary loss, but excludes directional loss and FiLM conditioning.}
    \label{fig:ablation}
\end{figure}

We conduct ablation studies to evaluate the contributions of individual design components in the collaborative VLA model. The results are presented in~\cref{fig:ablation}.

Several insights can be drawn from the experiments:

\begin{itemize}
    \item \textbf{Action post-processing} is the most critical component, yielding the largest improvement across all metrics.
    \item \textbf{Auxiliary loss} on hand pose provides consistent, though modest, gains—highlighting the benefit of including human-pose priors.
    \item \textbf{Directional loss} consistently reduces performance across metrics, suggesting it may overly constrain the learning dynamics.
    \item \textbf{FiLM conditioning} improves performance on low-dimensional objectives (e.g., L2 and PCA losses) but appears detrimental for other loss types.
\end{itemize}

\subsection{Auxiliary Predictions and Trainer Overfitting}
\label{auxiliary_predictions}

When we evaluate the model in real world set ups, we discovered an interesting fact: when trained on data collected from one specific collaborator, the model accurately interprets their intentions during inference. However, when interacting with a different person, it fails to adapt and instead reverts to a fixed routine—behavior as if no meaningful commands were received.

We named this phenomenon as \textbf{trainer overfitting}: the model becomes overly specialized to the behavior of a single demonstrator. This overfitting is also common in intelligent creatures, for example, dogs only follow the commands of their owners~\cite{merola2012dogs}. To further analyze this phenomenon, we conducted an experiment that uses auxiliary loss to quantify \textbf{trainer overfitting}.

\begin{figure}
    \centering
    \includegraphics[width=0.9\linewidth]{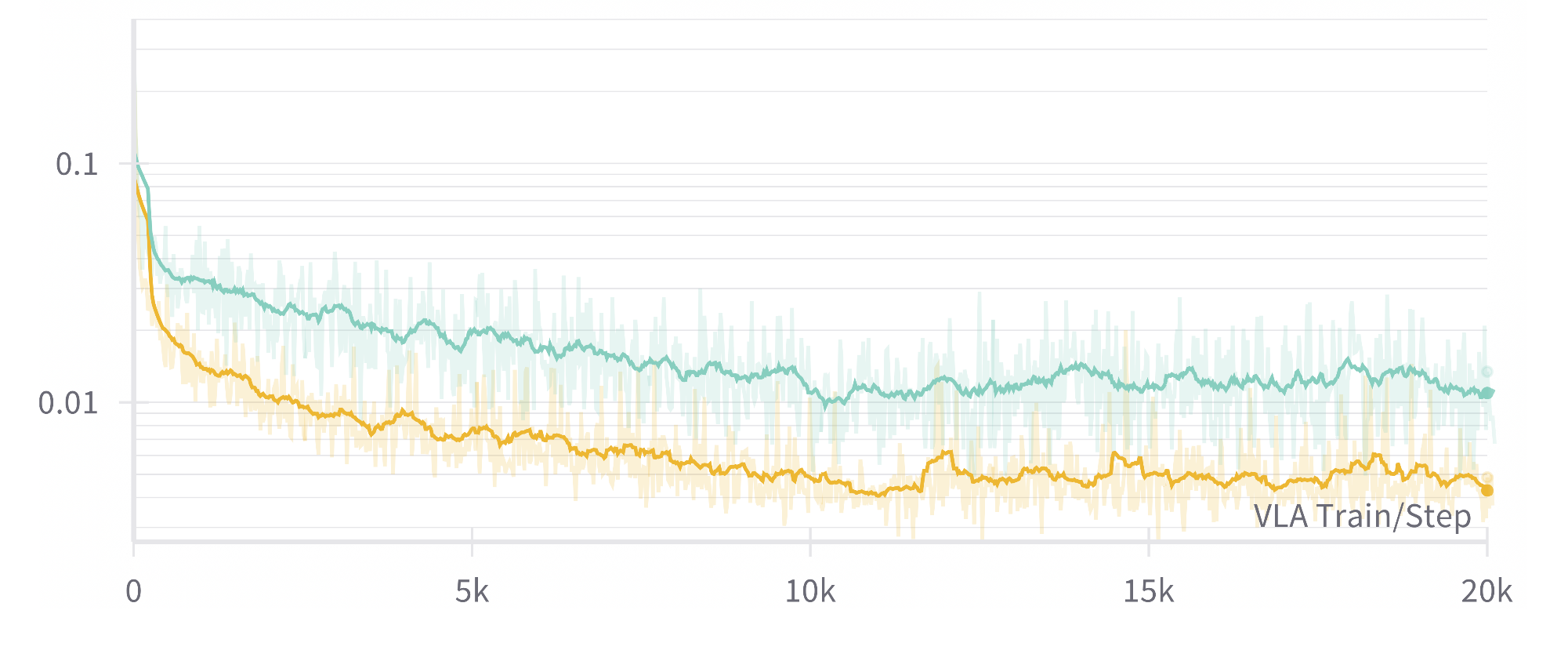}
    \caption{Auxiliary loss when evaluated on same-hand vs. different-hand collaborators. Orange curve: same hand as training. Blue curve: different hand.}
    \label{fig:two-hand}
\end{figure}

We trained the model with human collaborator A and tested the model on data from both collaborator A and collaborator B, and plotted the loss curve in~\cref{fig:two-hand}. The elevated loss confirms that the model fails to generalize across different collaborators.

This finding opens new possible research directions on how to reduce \textbf{trainer overfitting} in collaborative robots, or maybe, service dogs.

\subsection{Real World Evaluations}

\begin{figure}
    \centering
    \includegraphics[width=0.96\linewidth]{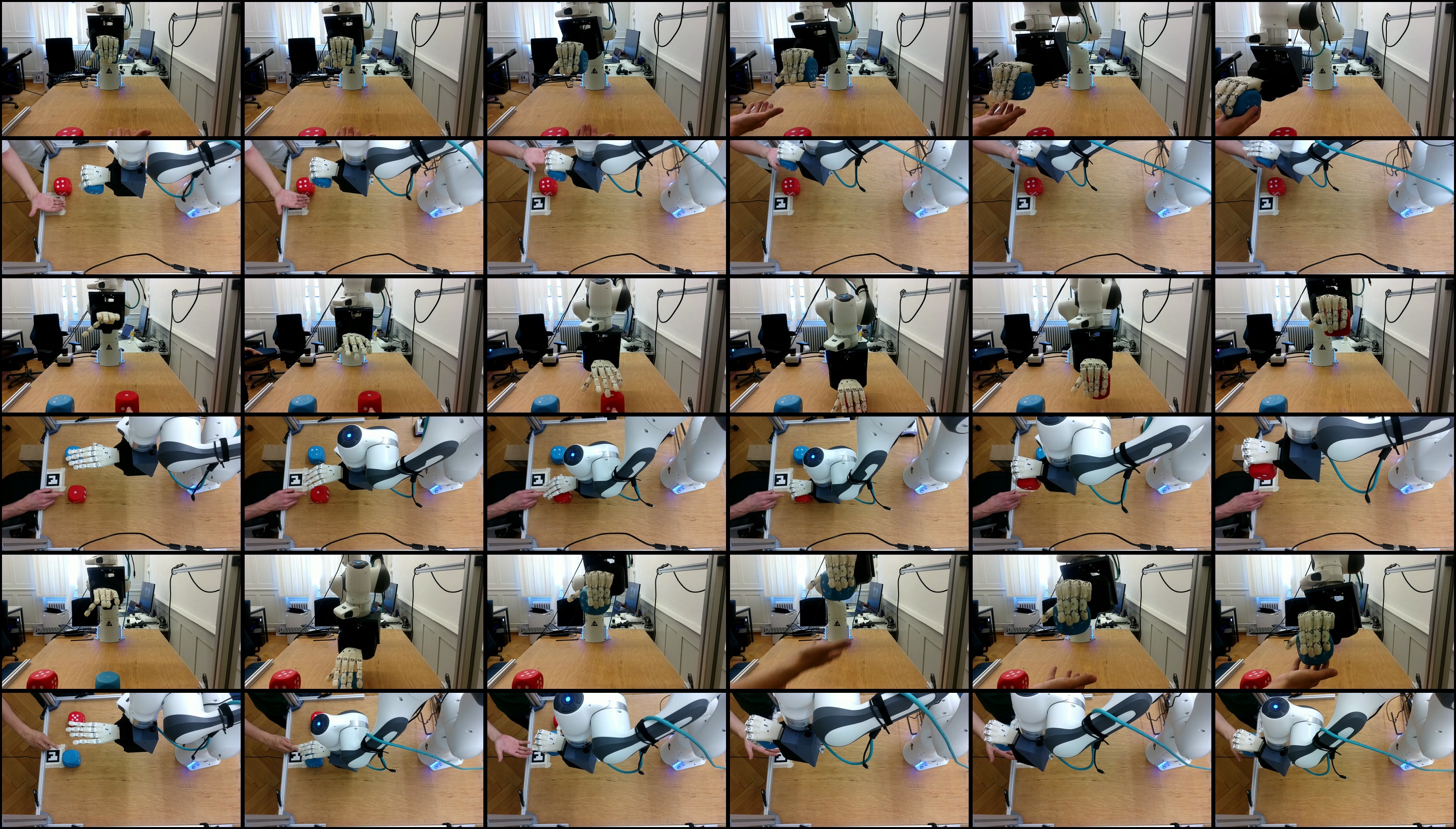}
    \caption{The snapshots of the robot carrying out successful real-world inference. The rows from top to bottom are the front and top view of task \textit{pass cube}, front and top view of task \textit{pick up cube} and front and top view of the combined long-horizon task. Each sequence is executed from left to right.}
    \label{fig:frame-matrix}
\end{figure}

We carried out a real-world evaluation on the model. The snap shots of real-world executions are included in~\cref{fig:frame-matrix}.

With the inference pipeline introduced in~\cref{inference_pipeline}, we are able to test the model in real-time interactively. Due to time constraints, we are not able to roll out enough trials for all the tasks to report success rate for each ablation on each task. In addition, during testing, the human collaborator is different from the trained data collected, thus influencing the overall performance of the real world evaluation due to the \textbf{trainer overfitting} phenomenon we found in~\cref{auxiliary_predictions}.

The model completed the combined long-horizon task (pick up then pass the cube) successfully once, in a total number of 10 trials. In all the other 9 cases, the model failed to recognize the cube that the human collaborator is pointing to, thus could not pick up the cube.

We believe that with more diverse training data collected, the model can do better in real-world scenarios.

\section{Conclusion}

In this project, we present a method for enabling robots to collaborate with humans through tacit understanding instead of relying solely on language-based prompting. We adapt a pre-trained Vision-Language-Action (VLA) model to collaborative tasks by introducing several architectural modifications, including FiLM conditioning, auxiliary hand-pose prediction, and action-space post-processing. These modifications improve the model’s ability to perceive, interpret, and respond to human intentions.

To demonstrate the effectiveness of our approach, we constructed a real-world collaborative dataset and designed two representative tasks: \textit{pick up cube} and \textit{pass cube}. By combining these tasks, we further demonstrated the model’s ability to handle long-horizon interactions. Our analysis shows that the proposed architectural changes are effective, and that the model can be trained efficiently on modest-scale hardware.

Our findings suggest that large VLA models can be effectively adapted to physical collaboration tasks when equipped with the appropriate inductive biases. This approach opens the door to more intuitive and efficient forms of human-robot interaction.






\section{Challenges and Future Work}

While the collaborative VLA framework demonstrates a promising direction for enabling intuitive human-robot interaction, several challenges remain.

A key issue is \textit{trainer overfitting} (see~\cref{auxiliary_predictions}), where the model becomes overly specialized to a single human demonstrator. This limits generalization across different users. Although this issue may diminish with larger-scale training and more diverse collaborators, it remains a significant limitation in the current system.

Another critical challenge is latency. Collaborative interaction is highly sensitive to response time, requiring the robot to react rapidly to human motions. Our current inference pipeline exhibits a latency of approximately 0.3 seconds—marginally acceptable in real-world settings. Reducing this latency, potentially through techniques such as temporal ensembling~\cite{zhao2023learning}, is essential for improving the fluidity of interaction.

Finally, the system's high-level planning is currently rule-based, limiting adaptability in dynamic environments. More flexible approaches—such as embodied chain-of-thought reasoning~\cite{zawalski2024robotic}—may offer improved performance in task sequencing and long-horizon planning.

Addressing these challenges will be crucial for deploying collaborative VLA systems in more complex, open-ended scenarios.

\clearpage

{
\bibliographystyle{plainnat}
\bibliography{reference}

\begin{thebibliography}{33}
\providecommand{\natexlab}[1]{#1}
\providecommand{\url}[1]{\texttt{#1}}
\expandafter\ifx\csname urlstyle\endcsname\relax
  \providecommand{\doi}[1]{doi: #1}\else
  \providecommand{\doi}{doi: \begingroup \urlstyle{rm}\Url}\fi

\bibitem[Achiam et~al.(2023)Achiam, Adler, Agarwal, Ahmad, Akkaya, Aleman, Almeida, Altenschmidt, Altman, Anadkat, et~al.]{achiam2023gpt}
Josh Achiam, Steven Adler, Sandhini Agarwal, Lama Ahmad, Ilge Akkaya, Florencia~Leoni Aleman, Diogo Almeida, Janko Altenschmidt, Sam Altman, Shyamal Anadkat, et~al.
\newblock Gpt-4 technical report.
\newblock \emph{arXiv preprint arXiv:2303.08774}, 2023.

\bibitem[Adam et~al.(2014)]{adam2014method}
Kingma DP Ba~J Adam et~al.
\newblock A method for stochastic optimization.
\newblock \emph{arXiv preprint arXiv:1412.6980}, 1412\penalty0 (6), 2014.

\bibitem[Anil et~al.(2023)Anil, Dai, Firat, Johnson, Lepikhin, Passos, Shakeri, Taropa, Bailey, Chen, et~al.]{anil2023palm}
Rohan Anil, Andrew~M Dai, Orhan Firat, Melvin Johnson, Dmitry Lepikhin, Alexandre Passos, Siamak Shakeri, Emanuel Taropa, Paige Bailey, Zhifeng Chen, et~al.
\newblock Palm 2 technical report.
\newblock \emph{arXiv preprint arXiv:2305.10403}, 2023.

\bibitem[Arora and Doshi(2021)]{arora2021survey}
Saurabh Arora and Prashant Doshi.
\newblock A survey of inverse reinforcement learning: Challenges, methods and progress.
\newblock \emph{Artificial Intelligence}, 297:\penalty0 103500, 2021.

\bibitem[Brohan et~al.(2022)Brohan, Brown, Carbajal, Chebotar, Dabis, Finn, Gopalakrishnan, Hausman, Herzog, Hsu, et~al.]{brohan2022rt}
Anthony Brohan, Noah Brown, Justice Carbajal, Yevgen Chebotar, Joseph Dabis, Chelsea Finn, Keerthana Gopalakrishnan, Karol Hausman, Alex Herzog, Jasmine Hsu, et~al.
\newblock Rt-1: Robotics transformer for real-world control at scale.
\newblock \emph{arXiv preprint arXiv:2212.06817}, 2022.

\bibitem[Brown et~al.(2020)Brown, Mann, Ryder, Subbiah, Kaplan, Dhariwal, Neelakantan, Shyam, Sastry, Askell, et~al.]{brown2020language}
Tom Brown, Benjamin Mann, Nick Ryder, Melanie Subbiah, Jared~D Kaplan, Prafulla Dhariwal, Arvind Neelakantan, Pranav Shyam, Girish Sastry, Amanda Askell, et~al.
\newblock Language models are few-shot learners.
\newblock \emph{Advances in neural information processing systems}, 33:\penalty0 1877--1901, 2020.

\bibitem[B{\"u}tepage et~al.(2020)B{\"u}tepage, Ghadirzadeh, {\"O}ztimur~Karadaǧ, Bj{\"o}rkman, and Kragic]{butepage2020imitating}
Judith B{\"u}tepage, Ali Ghadirzadeh, {\"O}zge {\"O}ztimur~Karadaǧ, M{\aa}rten Bj{\"o}rkman, and Danica Kragic.
\newblock Imitating by generating: Deep generative models for imitation of interactive tasks.
\newblock \emph{Frontiers in Robotics and AI}, 7:\penalty0 47, 2020.

\bibitem[Driess et~al.(2023)Driess, Xia, Sajjadi, Lynch, Chowdhery, Wahid, Tompson, Vuong, Yu, Huang, et~al.]{driess2023palm}
Danny Driess, Fei Xia, Mehdi~SM Sajjadi, Corey Lynch, Aakanksha Chowdhery, Ayzaan Wahid, Jonathan Tompson, Quan Vuong, Tianhe Yu, Wenlong Huang, et~al.
\newblock Palm-e: An embodied multimodal language model.
\newblock \emph{arXiv preprint arXiv:2303.03378}, 2023.

\bibitem[{Google AI Edge}(2024)]{mediapipe2024}
{Google AI Edge}.
\newblock Mediapipe.
\newblock \url{https://github.com/google-ai-edge/mediapipe}, 2024.
\newblock Accessed: 2025-07-13.

\bibitem[Huang et~al.(2023)Huang, Wang, Zhang, Li, Wu, and Fei-Fei]{huang2023voxposer}
Wenlong Huang, Chen Wang, Ruohan Zhang, Yunzhu Li, Jiajun Wu, and Li~Fei-Fei.
\newblock Voxposer: Composable 3d value maps for robotic manipulation with language models.
\newblock \emph{arXiv preprint arXiv:2307.05973}, 2023.

\bibitem[Huang et~al.(2018)Huang, Silv{\'e}rio, Rozo, and Caldwell]{huang2018generalized}
Yanlong Huang, Joao Silv{\'e}rio, Leonel Rozo, and Darwin~G Caldwell.
\newblock Generalized task-parameterized skill learning.
\newblock In \emph{2018 IEEE international conference on robotics and automation (ICRA)}, pages 5667--5474. IEEE, 2018.

\bibitem[Ji et~al.(2024)Ji, Zhang, Tang, Zheng, Liu, Zhao, and Li]{ji2024foundation}
Yuchen Ji, Zequn Zhang, Dunbing Tang, Yi~Zheng, Changchun Liu, Zhen Zhao, and Xinghui Li.
\newblock Foundation models assist in human--robot collaboration assembly.
\newblock \emph{Scientific Reports}, 14\penalty0 (1):\penalty0 24828, 2024.

\bibitem[Kim et~al.(2024)Kim, Pertsch, Karamcheti, Xiao, Balakrishna, Nair, Rafailov, Foster, Lam, Sanketi, et~al.]{kim2024openvla}
Moo~Jin Kim, Karl Pertsch, Siddharth Karamcheti, Ted Xiao, Ashwin Balakrishna, Suraj Nair, Rafael Rafailov, Ethan Foster, Grace Lam, Pannag Sanketi, et~al.
\newblock Openvla: An open-source vision-language-action model.
\newblock \emph{arXiv preprint arXiv:2406.09246}, 2024.

\bibitem[Kim et~al.(2025)Kim, Finn, and Liang]{kim2025fine}
Moo~Jin Kim, Chelsea Finn, and Percy Liang.
\newblock Fine-tuning vision-language-action models: Optimizing speed and success.
\newblock \emph{arXiv preprint arXiv:2502.19645}, 2025.

\bibitem[Lin et~al.(2023)Lin, Agia, Migimatsu, Pavone, and Bohg]{lin2023text2motion}
Kevin Lin, Christopher Agia, Toki Migimatsu, Marco Pavone, and Jeannette Bohg.
\newblock Text2motion: From natural language instructions to feasible plans.
\newblock \emph{Autonomous Robots}, 47\penalty0 (8):\penalty0 1345--1365, 2023.

\bibitem[Liu et~al.(2023)Liu, Li, Wu, and Lee]{liu2023visual}
Haotian Liu, Chunyuan Li, Qingyang Wu, and Yong~Jae Lee.
\newblock Visual instruction tuning.
\newblock \emph{Advances in neural information processing systems}, 36:\penalty0 34892--34916, 2023.

\bibitem[Liu et~al.(2024)Liu, Li, Li, and Lee]{liu2024improved}
Haotian Liu, Chunyuan Li, Yuheng Li, and Yong~Jae Lee.
\newblock Improved baselines with visual instruction tuning.
\newblock In \emph{Proceedings of the IEEE/CVF Conference on Computer Vision and Pattern Recognition}, pages 26296--26306, 2024.

\bibitem[Merola et~al.(2012)Merola, Prato-Previde, and Marshall-Pescini]{merola2012dogs}
Isabella Merola, Emanuela Prato-Previde, and Sarah Marshall-Pescini.
\newblock Dogs' social referencing towards owners and strangers.
\newblock \emph{PloS one}, 7\penalty0 (10):\penalty0 e47653, 2012.

\bibitem[Oquab et~al.(2023)Oquab, Darcet, Moutakanni, Vo, Szafraniec, Khalidov, Fernandez, Haziza, Massa, El-Nouby, et~al.]{oquab2023dinov2}
Maxime Oquab, Timoth{\'e}e Darcet, Th{\'e}o Moutakanni, Huy Vo, Marc Szafraniec, Vasil Khalidov, Pierre Fernandez, Daniel Haziza, Francisco Massa, Alaaeldin El-Nouby, et~al.
\newblock Dinov2: Learning robust visual features without supervision.
\newblock \emph{arXiv preprint arXiv:2304.07193}, 2023.

\bibitem[O’Neill et~al.(2024)O’Neill, Rehman, Maddukuri, Gupta, Padalkar, Lee, Pooley, Gupta, Mandlekar, Jain, et~al.]{o2024open}
Abby O’Neill, Abdul Rehman, Abhiram Maddukuri, Abhishek Gupta, Abhishek Padalkar, Abraham Lee, Acorn Pooley, Agrim Gupta, Ajay Mandlekar, Ajinkya Jain, et~al.
\newblock Open x-embodiment: Robotic learning datasets and rt-x models: Open x-embodiment collaboration 0.
\newblock In \emph{2024 IEEE International Conference on Robotics and Automation (ICRA)}, pages 6892--6903. IEEE, 2024.

\bibitem[Perez et~al.(2018)Perez, Strub, De~Vries, Dumoulin, and Courville]{perez2018film}
Ethan Perez, Florian Strub, Harm De~Vries, Vincent Dumoulin, and Aaron Courville.
\newblock Film: Visual reasoning with a general conditioning layer.
\newblock In \emph{Proceedings of the AAAI conference on artificial intelligence}, volume~32, 2018.

\bibitem[Ross et~al.(2011)Ross, Gordon, and Bagnell]{ross2011reduction}
St{\'e}phane Ross, Geoffrey Gordon, and Drew Bagnell.
\newblock A reduction of imitation learning and structured prediction to no-regret online learning.
\newblock In \emph{Proceedings of the fourteenth international conference on artificial intelligence and statistics}, pages 627--635. JMLR Workshop and Conference Proceedings, 2011.

\bibitem[Roveda et~al.(2019)Roveda, Haghshenas, Caimmi, Pedrocchi, and Molinari~Tosatti]{roveda2019assisting}
Loris Roveda, Shaghayegh Haghshenas, Marco Caimmi, Nicola Pedrocchi, and Lorenzo Molinari~Tosatti.
\newblock Assisting operators in heavy industrial tasks: On the design of an optimized cooperative impedance fuzzy-controller with embedded safety rules.
\newblock \emph{Frontiers in Robotics and AI}, 6:\penalty0 75, 2019.

\bibitem[Team et~al.(2024)Team, Ghosh, Walke, Pertsch, Black, Mees, Dasari, Hejna, Kreiman, Xu, et~al.]{team2024octo}
Octo~Model Team, Dibya Ghosh, Homer Walke, Karl Pertsch, Kevin Black, Oier Mees, Sudeep Dasari, Joey Hejna, Tobias Kreiman, Charles Xu, et~al.
\newblock Octo: An open-source generalist robot policy.
\newblock \emph{arXiv preprint arXiv:2405.12213}, 2024.

\bibitem[Touvron et~al.(2023)Touvron, Martin, Stone, Albert, Almahairi, Babaei, Bashlykov, Batra, Bhargava, Bhosale, et~al.]{touvron2023llama}
Hugo Touvron, Louis Martin, Kevin Stone, Peter Albert, Amjad Almahairi, Yasmine Babaei, Nikolay Bashlykov, Soumya Batra, Prajjwal Bhargava, Shruti Bhosale, et~al.
\newblock Llama 2: Open foundation and fine-tuned chat models.
\newblock \emph{arXiv preprint arXiv:2307.09288}, 2023.

\bibitem[Wang et~al.(2024)Wang, Hasler, Tanneberg, Ocker, Joublin, Ceravola, Deigmoeller, and Gienger]{wang2024lami}
Chao Wang, Stephan Hasler, Daniel Tanneberg, Felix Ocker, Frank Joublin, Antonello Ceravola, Joerg Deigmoeller, and Michael Gienger.
\newblock Lami: Large language models for multi-modal human-robot interaction.
\newblock In \emph{Extended Abstracts of the CHI Conference on Human Factors in Computing Systems}, pages 1--10, 2024.

\bibitem[Wojtak et~al.(2021)Wojtak, Ferreira, Vicente, Louro, Bicho, and Erlhagen]{wojtak2021neural}
Weronika Wojtak, Flora Ferreira, Paulo Vicente, Lu{\'\i}s Louro, Estela Bicho, and Wolfram Erlhagen.
\newblock A neural integrator model for planning and value-based decision making of a robotics assistant.
\newblock \emph{Neural Computing and Applications}, 33\penalty0 (8):\penalty0 3737--3756, 2021.

\bibitem[Yan et~al.(2019)Yan, Gao, Zhang, and Chang]{yan2019human}
Liang Yan, Xiaoshan Gao, Xiongjie Zhang, and Suokui Chang.
\newblock Human-robot collaboration by intention recognition using deep lstm neural network.
\newblock In \emph{2019 IEEE 8th International Conference on Fluid Power and Mechatronics (FPM)}, pages 1390--1396. IEEE, 2019.

\bibitem[Zare et~al.(2024)Zare, Kebria, Khosravi, and Nahavandi]{zare2024survey}
Maryam Zare, Parham~M Kebria, Abbas Khosravi, and Saeid Nahavandi.
\newblock A survey of imitation learning: Algorithms, recent developments, and challenges.
\newblock \emph{IEEE Transactions on Cybernetics}, 2024.

\bibitem[Zawalski et~al.(2024)Zawalski, Chen, Pertsch, Mees, Finn, and Levine]{zawalski2024robotic}
Micha{\l} Zawalski, William Chen, Karl Pertsch, Oier Mees, Chelsea Finn, and Sergey Levine.
\newblock Robotic control via embodied chain-of-thought reasoning.
\newblock \emph{arXiv preprint arXiv:2407.08693}, 2024.

\bibitem[Zhai et~al.(2023)Zhai, Mustafa, Kolesnikov, and Beyer]{zhai2023sigmoid}
Xiaohua Zhai, Basil Mustafa, Alexander Kolesnikov, and Lucas Beyer.
\newblock Sigmoid loss for language image pre-training.
\newblock In \emph{Proceedings of the IEEE/CVF international conference on computer vision}, pages 11975--11986, 2023.

\bibitem[Zhang et~al.(2020)Zhang, Liu, Chang, Wang, and Gao]{zhang2020recurrent}
Jianjing Zhang, Hongyi Liu, Qing Chang, Lihui Wang, and Robert~X Gao.
\newblock Recurrent neural network for motion trajectory prediction in human-robot collaborative assembly.
\newblock \emph{CIRP annals}, 69\penalty0 (1):\penalty0 9--12, 2020.

\bibitem[Zhao et~al.(2023)Zhao, Kumar, Levine, and Finn]{zhao2023learning}
Tony~Z Zhao, Vikash Kumar, Sergey Levine, and Chelsea Finn.
\newblock Learning fine-grained bimanual manipulation with low-cost hardware.
\newblock \emph{arXiv preprint arXiv:2304.13705}, 2023.

\end{thebibliography}
}

\end{document}